\documentclass[11pt]{article}

\usepackage{algorithm}
\usepackage{algorithmic}
\makeatletter
\newcommand{\blfootnote}[1]{%
  \begingroup
  \renewcommand\@makefntext[1]{##1}
  \footnotetext{#1}%
  \endgroup
}
\makeatother
\usepackage{booktabs,multirow,graphicx,amssymb} 
\usepackage{sidecap}
\usepackage{graphicx}
\usepackage{rotating}
\usepackage{agenticlearning-xelatex}

\begin{document}

\shorttitle{AdaJEPA}
\shortauthor{Wang \etal}

\title{AdaJEPA: An Adaptive Latent World Model}
\author{
  Ying Wang$^{12}$, Oumayma Bounou$^{1}$, Yann LeCun$^{*12}$, Mengye Ren$^{*1}$ \\
  $^{1}$New York University $^{2}$AMI Labs\\
  \texttt{\{yw3076,ob2184,yann.lecun,mengye\}@nyu.edu}\\
  \url{https://agenticlearning.ai/adajepa}
}
\date{}
\blfootnote{* Equal advising.}
\maketitle

\begin{abstract}
Latent world models enable planning from high-dimensional observations by predicting future states in a compact latent space. However, these models are typically kept frozen at test time: when their predictions become inaccurate, planning can fail, especially under test-time distribution shift. To address this, we propose AdaJEPA, an adaptive latent world model that performs test-time adaptation within the closed loop of model predictive control (MPC). After training, AdaJEPA plans and executes the first action chunk, uses the observed next-state transition as a self-supervised adaptation signal, and replans with the updated model. This closed-loop update continuously recalibrates the world model without additional expert demonstrations. Across a range of goal-reaching tasks, AdaJEPA substantially improves planning success with as few as one gradient step per MPC replanning step.

\end{abstract}
\section{Introduction}

The long-standing goal of latent world models is to capture environment dynamics in a compact latent space that enables efficient prediction and planning with better generalization. Joint-Embedding Predictive Architectures (JEPAs) have emerged as a powerful world model paradigm that jointly learns an encoder and a predictor by optimizing a latent prediction objective on reward-free offline trajectories~\citep{jepa, sobal2025learning, assran2025vjepa2selfsupervisedvideo, dinowm}. Within this framework, the planning task is defined at test time and is often performed with model predictive control (MPC)~\citep{mpc1989,kouvaritakis2016model}, which repeatedly rolls out the model forward, optimizes a short-horizon action sequence, executes the first (or first few) action(s), and replans from the next observation. This combination of world models with MPC has become a standard recipe for goal-conditioned control and planning.

Yet, the world model used for planning is typically frozen after training. When its predictions are inaccurate, MPC optimizes the wrong objective: actions that appear effective in latent rollouts may fail under the true system dynamics. Even small one-step errors can compound over the planning horizon, causing actions that appear effective in latent imagination to fail in the real environment. This issue is amplified under test-time distribution shift: visual changes such as noise, lighting, or background distractors can misalign the encoder, while changes in friction, mass, or contact dynamics can misalign the latent predictor. As a result, even high-capacity world models can suffer substantial performance degradation when small changes occur in test environments~\citep{dinowm,toso2026learning}, which hinders the application of world models in the real world. This suggests that \emph{a world model should not remain fixed after training, but continue to improve from the experience encountered during deployment}. This intuition is well-grounded in biological systems, where sensorimotor adaptation relies on cerebellar mechanisms to adjust behavior under changing inputs and dynamics~\citep{ shadmehr1994adaptive,wolpert1998internal,bastian2006learning,shadmehr2010error}. We humans also continually update internal world models with new experience, and these updates shape subsequent decisions~\citep{craik1943nature,Glscher2010StatesVR,nassar2010approximately,daw2011model}.

\begin{figure*}[t]
  \centering
    \centering
    \begin{subfigure}{0.64\textwidth}
        \centering
        \includegraphics[width=\linewidth]{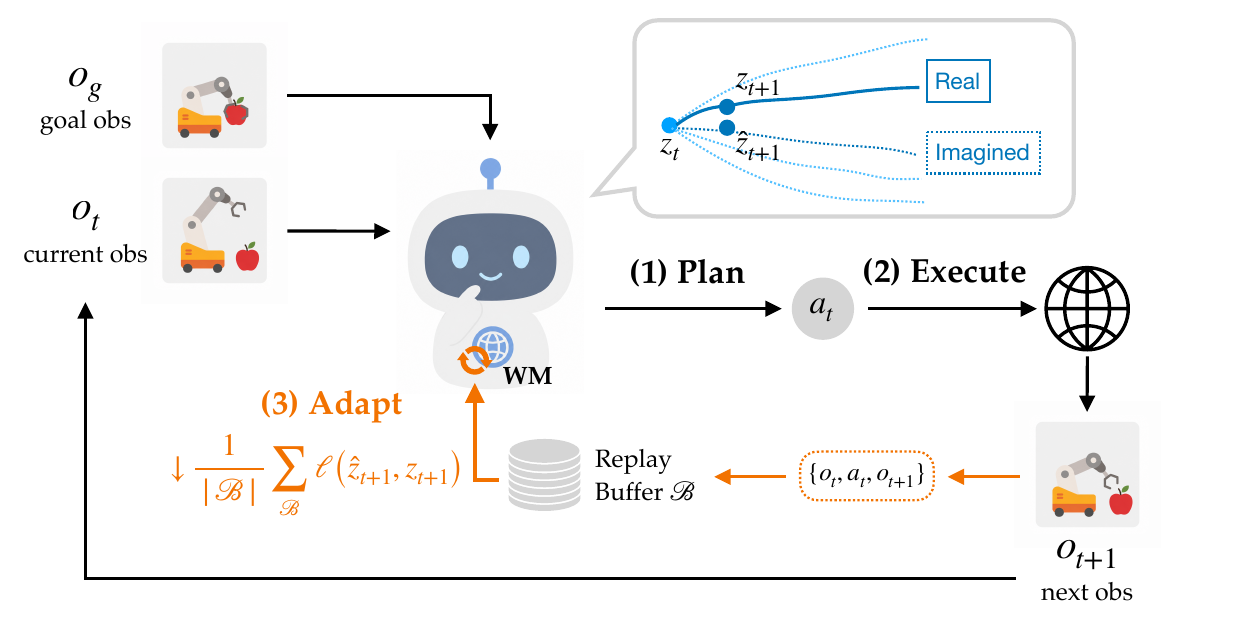}
        \caption{AdaJEPA: Plan–Act–Adapt–Replan Loop}
        \label{fig:main_loop}
    \end{subfigure}
    \hfill
    \begin{subfigure}{0.35\textwidth}
        \centering
        \includegraphics[width=\linewidth]{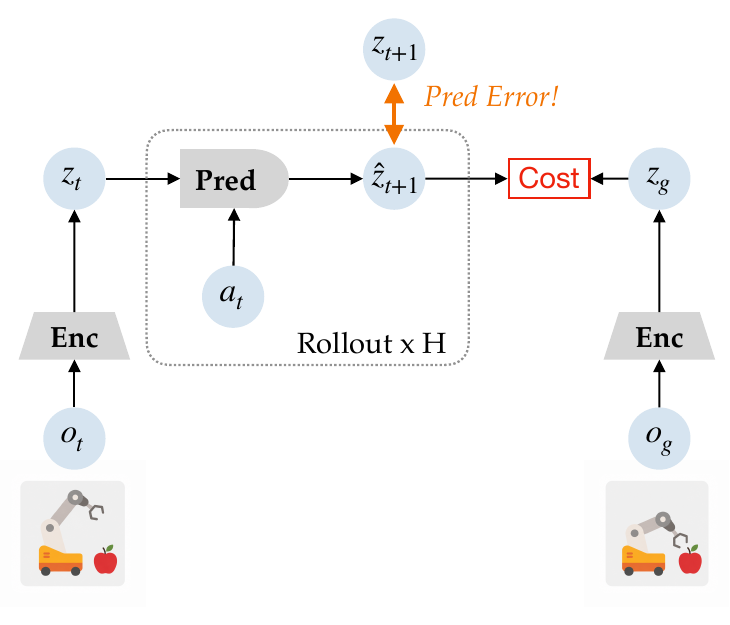}
        \caption{World Model for Planning}
        \label{fig:main_plan}
    \end{subfigure}
  \caption{AdaJEPA performs test-time adaptation during closed-loop MPC. At each MPC step, we plan with the current model, execute the first action $a_t$, collect observation $o_{t+1}$ from the environment, and update the model to minimize the prediction error on the newly observed transition $\{o_t,a_t,o_{t+1}\}$ before replanning. This yields a simple plan–execute–adapt–replan loop that continually recalibrates the model to transitions encountered in the current environment.
  }
  \label{fig:main}
\end{figure*}

To adapt world models during deployment, we propose \textbf{AdaJEPA}, a test-time adaptation framework that operates within the closed loop of MPC. Rather than treating the model as fixed after training, AdaJEPA uses the transition observed after each executed action as a self-supervised signal to adapt the world model before the next replan~(\Cref{fig:main}). This makes learning and planning tightly coupled: the model optimizes actions using its predictions, the consequence of the actions provides the adaptation signal, and the adapted model improves prediction and thus planning for the next iteration. The adaptation is both sample- and compute-efficient, as each update modifies only a small subset of parameters and can be performed with a single gradient step on the latest transition of the current episode. AdaJEPA significantly outperforms the frozen model across in-distribution and out-of-distribution goal-reaching tasks, with particularly strong gains when training data is limited.

\section{Related Work}

\paragraph{JEPA world models.}
Joint-Embedding Predictive Architectures (JEPAs) encode high-dimensional sensory observations into a compact latent space and learn dynamics within it. They consist of an encoder and a predictor which are typically trained jointly by minimizing a latent prediction objective over reward-free offline trajectories~\citep{jepa, dinowm,sobal2025learning, assran2025vjepa2selfsupervisedvideo,dexwm,wang2026temporal_straightening, maes2026leworldmodel,zhang2026hierarchical}. Once trained, these models are paired with planners such as MPC to solve goal-conditioned tasks. However, all of the aforementioned world models are kept frozen at test time, leading to potential performance degradation under test-time distribution shift. Closest in motivation, \cite{parthasarathy2025closing} aim to reduce the train-test gap through data synthesis during training, but their model still remains fixed during planning and is evaluated only on in-distribution environments. To the best of our knowledge, we are the first to adapt a JEPA world model during planning.

\paragraph{Test-time training and adaptation.}
Test-time training/adaptation (TTT/TTA) updates a pretrained model using test data to reduce test-time distribution shift. Unlike supervised finetuning on a labeled target dataset, TTT/TTA typically optimizes an unsupervised or self-supervised objective computed from the test input itself. Early applications mainly focus on image classification with common objectives including auxiliary tasks~\citep{sun2020ttt}, entropy minimization~\citep{wang2021tent,niu2022eata,wang2022cotta,niu2023sar}, augmentation-based consistency~\citep{zhang2022memo}, and masked reconstruction~\citep{gandelsman2022tttmae}. 
The same principle has since been explored in a wide range of domains, including in-context learning~\citep{akyurek2024tttllm} and embodied decision making~\citep{hong2026reflective} with LLMs, prompt tuning of vision-language models~\citep{yoon2024ctpt}, and streaming videos~\citep{wang2023videottt}. Our work applies test-time adaptation to latent planning with world models: each execution leads to the next observation, which serves as a self-supervised prediction target for adaptation. 

\paragraph{Adaptation in planning and control.}
The idea of updating predictive models during decision making dates back to adaptive control: for example, self-adapting IDCOM identifies a low-dimensional input--output process model online and uses it to retune the predictive controller~\citep{foigel1979selfadaptingidcom}. Similarly, online model-based RL updates dynamics models from interaction, but these updates are usually coupled to policy or value learning~\citep{dyna,thrun1990planning,hansen2022temporal,hansen2023td}. Recently, several world models also focus on adapting pretrained predictors, but typically require target-domain finetuning data, additional online rollouts, or outer self-improvement loops~\citep{wang2025adawm,gao2025adaworld,lanier2025redraw,luoself}. AdaJEPA instead adapts a pretrained world model inside closed-loop MPC: each executed transition provides a self-supervised latent prediction target, and the updated world model is reused at the next replan, \emph{without reward labels, expert labels, or a separate data-collection phase}.

\section{AdaJEPA: An Adaptive Latent World Model}
\label{sec:method}

Unlike existing world models that are kept frozen during planning, AdaJEPA performs test-time adaptation during closed-loop MPC. At test time, we plan with the current model, execute the first action, and update the model using the newly observed transition before replanning. This yields a simple plan--execute--adapt--replan loop that continually recalibrates the model to transitions encountered during planning. 

\subsection{Background: JEPA World Models}

We consider trajectories of high-dimensional observations $o_t \in \mathbb{R}^{n_o}$ and actions $a_t \in \mathbb{R}^{n_a}$. 
A latent world model consists of a sensory encoder $\mathcal{E}^s_\phi$, an action encoder $\mathcal{E}^a_\psi$, and a predictor $f_\theta$:
\begin{equation}
z_t=\mathcal{E}^s_\phi(o_t), 
\qquad 
u_t=\mathcal{E}^a_\psi(a_t),
\qquad
\hat z_{t+1}=f_\theta(z_t,u_t).
\end{equation}
$f_\theta$ may be conditioned on a short history of latent states and action embeddings, and we use the one-step notation for simplicity. The encoder and predictor are jointly trained on a reward-free offline transition data $\mathcal{D}_{\rm off}=\{(o_t,a_t,o_{t+1})\}$ by predicting future latent targets rather than reconstructing pixels. A generic JEPA-style prediction objective is
\begin{equation}
\mathcal{L}_{\rm pred}
=
\frac{1}{K} \sum_{k=1}^{K} \ell(\hat z_{t+k}, z_{t+k}),
\end{equation}
where $z_{t+k}$ is the target representation of $o_{t+k}$ and $\ell$ is a latent prediction loss such as MSE. 
Different JEPA instantiations may either use the stop-gradient operator~\citep{chen2020simsiam} on the target representation or include a regularization term~\citep{bardes2022vicregvarianceinvariancecovarianceregularizationselfsupervised, balestriero2025lejepa} in the training objective to prevent collapse.

After training, the world model can be used for goal-conditioned latent planning. Given a goal observation $o_g$ with latent representation $z_g=\mathcal{E}^s_\phi(o_g)$, MPC optimizes an action sequence by rolling out $f_\theta$ and minimizing a latent goal-reaching cost (\Cref{fig:main_plan}):
\begin{equation}
a^*_{t:t+H-1}
=
\arg\min_{a_{t:t+H-1}}
\sum_{k=1}^{H}
\alpha_k\, d(\hat z_{t+k},z_g),
\label{eq:mpc_objective}
\end{equation}
where $H$ is the planning horizon, $\alpha_k$ are temporal weights, and $d$ is typically the squared Euclidean distance, $d(\hat z,z_g)=\|\hat z-z_g\|_2^2$. The optimization in~\eqref{eq:mpc_objective} can be solved with either gradient-based planners or sampling-based methods such as CEM~\citep{rubinstein1997optimization}. Standard MPC executes the first action $a_t$, after which the agent observes the next observation $o_{t+1}$ and replans.

\subsection{Closed-Loop Plan-and-Adapt}
\label{sec:ada_loop}

A pretrained world model is never perfect. Prediction errors can arise from finite offline data and test-time distribution shifts. To minimize prediction errors at deployment and thus improve planning success, AdaJEPA continuously updates itself using the transitions caused by its own actions. 

\Cref{alg:adajepa} details the proposed \emph{plan-execute-adapt-replan} loop (visualized in \Cref{fig:main_loop}) for one episode. At each MPC replanning step, we plan with the current model, execute the first action, observe the resulting transition, perform a small number of self-supervised updates, and replan with the updated model. Note that each episode starts from the same pretrained model and maintains its own copy and buffer throughout the episode. We describe the key components of the test-time adaptation in the following. 

\paragraph{Online buffer.}
The buffer $\mathcal{B}$ stores recent transitions collected during MPC. After executing $a_t$ and observing $o_{t+1}$, we append $(o_t,a_t,o_{t+1})$ to $\mathcal{B}$. Because $\mathcal{B}$ can grow unbounded, we cap it to a fixed size $N$
at each iteration. We consider two strategies: (i) \texttt{recent-N} keeps only the most recent $N$ transitions, which focuses adaptation on the local observations and dynamics currently encountered, and (ii) \texttt{hard-N} keeps only the $N$ transitions that result in the largest prediction errors (comparison is in~\Cref{app:ablations}).

\paragraph{Adaptation loss.}
AdaJEPA uses the same self-supervised prediction signal at test time as in pretraining. For clarity, we assume one-step history and no frameskip. With the replay buffer $\mathcal{B}$, the adaptation loss is
\begin{equation}
\mathcal{L}_{\rm ada}(\mathcal{B})
=
\frac{1}{|\mathcal{B}|}
\sum_{(o_i,a_i,o_{i+1})\in\mathcal{B}}
\ell\!\left(
f_\theta\!\left(z_i,\mathcal{E}^a_\psi(a_i)\right),
\operatorname{sg}(z_{i+1})
\right),
\label{eq:ada_loss}
\end{equation}
where $z_i=\mathcal{E}^s_\phi(o_i)$, $\ell$ is a latent prediction loss, and $\operatorname{sg}(\cdot)$ denotes the stop-gradient operator. Here, we use stop-gradient as the default anti-collapse stabilizer during online adaptation, though it can be replaced by other methods depending on the pretrained world model. Removing stop-gradient while updating only the last layers of the encoder and predictor for one step gives similar planning performance, suggesting that the restricted online update already limits collapse. For longer histories, action chunks, and frameskips, we average the same loss over all valid prediction windows.

\begin{algorithm}[t]
\caption{AdaJEPA: Closed-Loop Plan-and-Adapt}
\label{alg:adajepa}
\small
\begin{algorithmic}[1]
\STATE \textbf{Input:} pretrained world model $(\mathcal{E}^s_\phi,\mathcal{E}^a_\psi,f_\theta)$, trainable parameters $\Omega$, goal observation $o_g$, planning horizon $H$, adaptation steps $U$, buffer size $N$, max steps $T$
\STATE Initialize buffer $\mathcal{B}\leftarrow\emptyset$; observe $o_0$
\FOR{$t=0,1,\ldots,T-1$}
    \STATE Plan with the current model: optimizes an action sequence by minimizing a latent goal-reaching cost (eq~\eqref{eq:mpc_objective})
    \STATE Execute the first action $a_t$ and observe $o_{t+1}$
    \STATE Add $(o_t,a_t,o_{t+1})$ to $\mathcal{B}$ and trim (if necessary) to keep $N$ transitions at maximum
    \FOR{$u=1,\ldots,U$}
        \STATE $\Omega \leftarrow \Omega-\eta\nabla_\Omega \mathcal{L}_{\rm ada}(\mathcal{B})$
    \ENDFOR
\ENDFOR
\end{algorithmic}
\end{algorithm}

\paragraph{Adapted parameters.}
Let $\Omega\subseteq\{\phi,\psi,\theta\}$ denote the parameters updated at test time. After each MPC step, AdaJEPA performs $U$ gradient updates,
\begin{equation}
\Omega \leftarrow \Omega-\eta\nabla_\Omega\mathcal{L}_{\rm ada}(\mathcal{B}).
\end{equation}
The adapted model is immediately reused for the next planning problem. In experiments, $\Omega$ can be restricted to a small subset of encoder or predictor parameters, making adaptation lightweight.

\vspace{-0.1in}
\section{Experiments}
\label{sec:experiments}

In this section, we evaluate the planning performance of our proposed AdaJEPA on different train--test distribution shifts and under different adaptation modes and design choices. Our results demonstrate that adaptation with only one GD step in each MPC replanning step can lead to strong and robust performance gains in all settings.

\subsection{Setup}

\begin{figure*}[t]
  \centering
  \includegraphics[width=\textwidth]{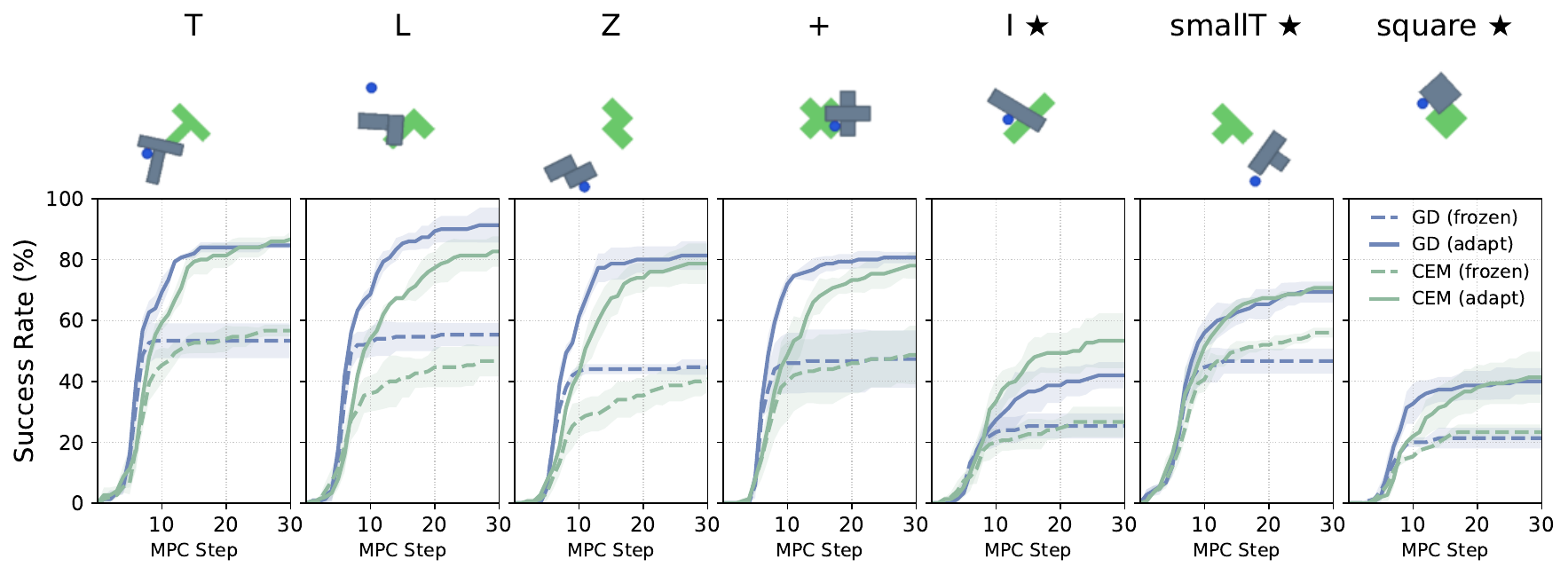}
  \vspace{0.08in}
  \includegraphics[width=\textwidth]{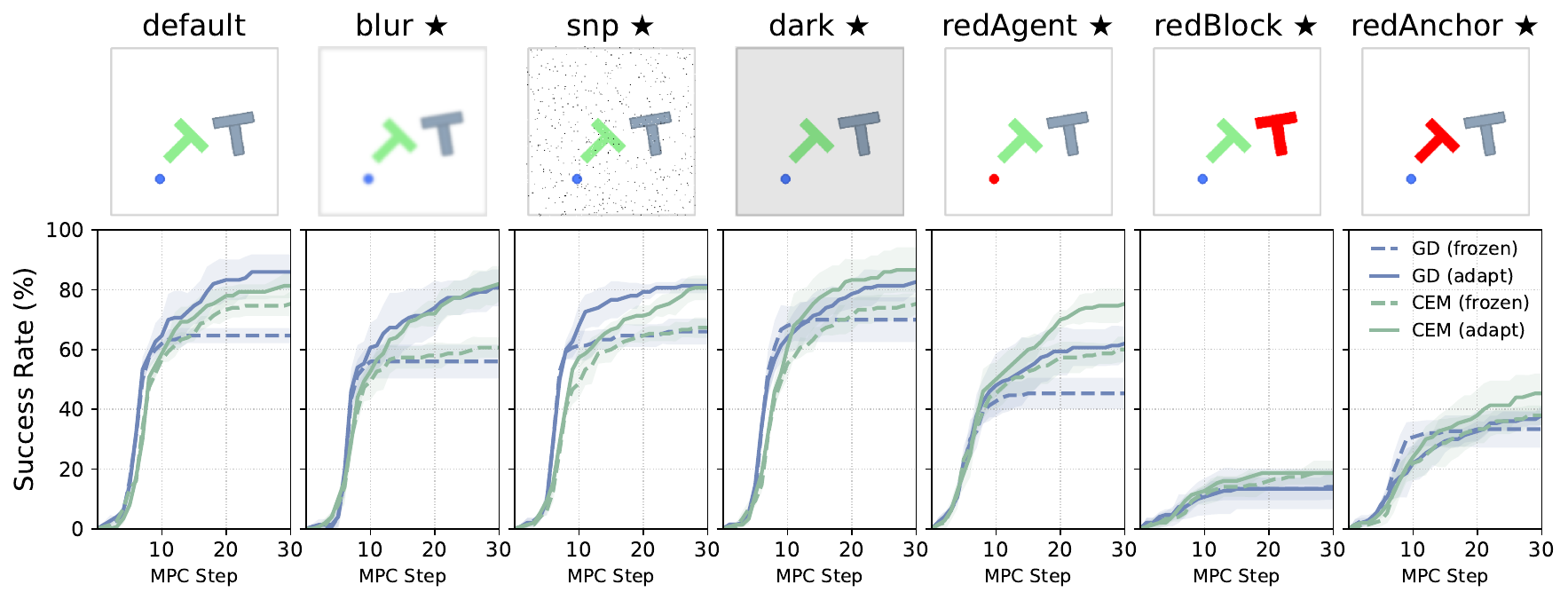}
  \vspace{-0.3in}
  \caption{
    Planning Success under Shape Shifts (top) and Visual Shifts (bottom). The $\bigstar$ denotes unseen shapes and configurations. AdaJEPA consistently improves planning success across all settings, using only a single adaptation step per MPC replanning step. We extend the maximum number of steps to 30 to show the increasing trend of planning success of AdaJEPA. Comparison between frozen and AdaJEPA planning trajectories is in~\Cref{app:viz}, showing AdaJEPA consistently decreases prediction loss and leads to better planning.
  }
  \label{fig:pushobj}
  \vspace{-0.2in}
\end{figure*}

\paragraph{Environments.} Our main results are conducted on the PushT~\citep{chi2024pusht} and PointMaze~\citep{fu2020d4rl} benchmarks (environment details in~\Cref{{app:env}}). We evaluate planning on unseen in-distribution trajectories, as well as on the following out-of-distribution variations:
\begin{itemize}
    \item \emph{Shape shifts} change the block in the PushT environment from a T to other shapes following the PushObj setup in~\citep{dinowm}, but keep the colors unchanged. We train on four shapes \{\texttt{T}, \texttt{L}, \texttt{Z}, \texttt{+}\} and test on both the training shapes and three held-out OOD shapes \{\texttt{I}, \texttt{smallT}, \texttt{square}\} that share the same contact dynamics but have unseen geometry. To ensure a reasonable contact ratio, we bias the training and testing data toward contact-rich interactions, forcing at least one contact in the test trajectory. Goals are sampled 25 steps away. Examples of shapes can be found in~\Cref{fig:pushobj} and more details on data generation are included in~\Cref{{app:pushobj}}.
    \item \emph{Visual shifts} apply per-frame corruptions to the original PushT observations by adding (i) Gaussian blur, (ii) salt-and-pepper (snp) noise, and (iii) dark lighting. Additionally, we also test the robustness to color shifts by changing (iv) the moving T from light gray to red, (v) the anchor T from light green to red, and (vi) the agent from blue to red. During training, the model is trained on the original PushT data but tested on these visual OOD settings. Goals are sampled 25 steps away. Examples of shapes can be found in~\Cref{fig:pushobj} and more details of data generation are included in~\Cref{{app:pushobj}}.
    \item \emph{Dynamics shifts} vary physics in the PointMaze-Medium environment. We change physics to (i) low mass (x0.2) which causes the agent to move faster under the same force, and (ii) high damping (x20) that causes velocity to decay faster. The model is trained on the default dynamics using the training data from~\cite{wang2026temporal_straightening} and tested on these OOD dynamics settings. Goals are sampled such that their grid distance from the start is larger than 3. More details are in~\Cref{app:pointmaze}.
    \item \emph{Layout shifts} vary the maze layouts in the PointMaze environment as proposed by~\citet{sobal2025learning, zhang2026hierarchical}. Maze layouts are randomly generated on an $8\times8$ grid with connected free space. We train on 25 maze layouts and evaluate on 5 held-out layouts. At test time, goals are sampled such that the shortest-path distance from the start is 3--5 cells, ensuring nontrivial but feasible navigation tasks. More details are in~\Cref{app:diversemaze}.
\end{itemize}

\paragraph{Plan-and-Adapt.}
We use receding-horizon MPC for all experiments, with either GD or CEM as the trajectory optimizer. Unless specified otherwise, AdaJEPA performs adaptation at every MPC replanning step by updating only the final layers of the visual encoder and predictor, while keeping all other parameters frozen. Each update consists of \emph{a single gradient step} using the same learning rates as training ($\eta_{\text{pred}} = 5 \times 10^{-4}$ and $\eta_{\text{enc}} = 10^{-5}$). We maintain a replay buffer containing the 5 most recent samples. At each MPC replanning step, only a single action chunk is executed, and the maximum number of MPC steps is set to 20. All reported results are averaged over three test-data seeds, with 50 episodes per seed.

\paragraph{Architectures.}
For each environment, we train a JEPA world model from scratch, following \citet{wang2026temporal_straightening}. The model consists of a ResNet encoder that produces global features and a transformer-based predictor. We apply stop-gradient to the target branch in the prediction loss to prevent collapse and use curvature regularization to encourage straighter latent trajectories to facilitate planning. Following prior work~\citep{dinowm,wang2026temporal_straightening}, action embeddings are concatenated with visual and proprioceptive embeddings before being passed to the predictor. We use a frameskip of 5 and a history window of 3. Implementation details are provided in \Cref{app:exp}. Note that AdaJEPA is \emph{agnostic to the underlying world model implementation}, and we show consistent improvements across model variants in \Cref{tab:pusht_ttt}.

\subsection{Results}

\begin{figure*}[t]
  \centering
  \setlength{\tabcolsep}{3pt}
  \small
  \captionof{table}{Planning Success under Dynamics and Layout Shifts for PointMaze. Note that low mass causes faster movement under same force and high damping causes velocity decay faster.}
  \label{tab:pointmaze_layout_dynamics}
\resizebox{\textwidth}{!}{%
\begin{tabular}{l *{8}{r@{\,{\tiny\,$\pm$\,}}l@{\;}l}}
    \toprule
    & \multicolumn{18}{c}{Dynamics Shift (PointMaze-Medium)}
    & \multicolumn{6}{c}{Layout Shift} \\
    \cmidrule(lr){2-19} \cmidrule(lr){20-25}
    & \multicolumn{6}{c}{default}
    & \multicolumn{6}{c}{low mass}
    & \multicolumn{6}{c}{high damping}
    & \multicolumn{6}{c}{Unseen Layouts} \\
    \cmidrule(lr){2-7}\cmidrule(lr){8-13}\cmidrule(lr){14-19}\cmidrule(lr){20-25}
    Adapt
    & \multicolumn{3}{c}{GD} & \multicolumn{3}{c}{CEM}
    & \multicolumn{3}{c}{GD} & \multicolumn{3}{c}{CEM}
    & \multicolumn{3}{c}{GD} & \multicolumn{3}{c}{CEM}
    & \multicolumn{3}{c}{GD} & \multicolumn{3}{c}{CEM} \\
    \cmidrule(lr){2-4}\cmidrule(lr){5-7}
    \cmidrule(lr){8-10}\cmidrule(lr){11-13}
    \cmidrule(lr){14-16}\cmidrule(lr){17-19}
    \cmidrule(lr){20-22}\cmidrule(lr){23-25}

    Frozen
    & 82.7 & {\tiny 6.8} &
    & 84.0 & {\tiny 3.3} &
    & 77.3 & {\tiny 8.2} &
    & 82.0 & {\tiny 2.8} &
    & 77.3 & {\tiny 5.0} &
    & 76.0 & {\tiny 2.8} &
    & 53.3 & {\tiny 8.2} &
    & 49.3 & {\tiny 6.2} &
    \\

    \texttt{predlast + enclast}
    & 83.3 & {\tiny 6.6} & {\scriptsize\textcolor{LabBlue}{$\uparrow$0.7}}
    & 83.3 & {\tiny 3.4} & {\scriptsize\textcolor{LabRed}{$\downarrow$0.7}}
    & 80.0 & {\tiny 3.3} & {\scriptsize\textcolor{LabBlue}{$\uparrow$2.7}}
    & 86.7 & {\tiny 2.5} & {\scriptsize\textcolor{LabBlue}{$\uparrow$4.7}}
    & 77.3 & {\tiny 10.5} &
    & 78.7 & {\tiny 3.4} & {\scriptsize\textcolor{LabBlue}{$\uparrow$2.7}}
    & 66.0 & {\tiny 7.1} & {\scriptsize\textcolor{LabBlue}{$\uparrow$12.7}}
    & 55.3 & {\tiny 5.0} & {\scriptsize\textcolor{LabBlue}{$\uparrow$6.0}}
    \\

    \texttt{predfirst + enclast}
    & 84.0 & {\tiny 1.6} & {\scriptsize\textcolor{LabBlue}{$\uparrow$1.3}}
    & 84.0 & {\tiny 4.3} &
    & 82.0 & {\tiny 1.6} & {\scriptsize\textcolor{LabBlue}{$\uparrow$4.7}}
    & 82.7 & {\tiny 3.4} & {\scriptsize\textcolor{LabBlue}{$\uparrow$0.7}}
    & 78.7 & {\tiny 4.7} & {\scriptsize\textcolor{LabBlue}{$\uparrow$1.3}}
    & 82.0 & {\tiny 3.3} & {\scriptsize\textcolor{LabBlue}{$\uparrow$6.0}}
    & 78.7 & {\tiny 5.0} & {\scriptsize\textcolor{LabBlue}{$\uparrow$25.3}}
    & 70.7 & {\tiny 3.8} & {\scriptsize\textcolor{LabBlue}{$\uparrow$21.3}}
    \\

    \bottomrule
\end{tabular}}

  \vspace{0.1in}
  \centering
  \setcounter{subfigure}{0}%
  \renewcommand{\thesubfigure}{\ifcase\value{subfigure}\or a1\or a2\or a3\or b1\or b2\or b3\fi}%
  \captionsetup{type=figure}
  \begin{subfigure}{0.16\textwidth}
    \includegraphics[width=\linewidth]{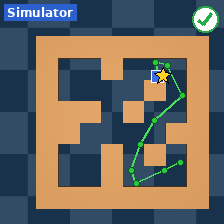}
    \caption{Default Dyn. \\ frozen $\rightarrow$ success}
  \end{subfigure}
  \begin{subfigure}{0.16\textwidth}
    \includegraphics[width=\linewidth]{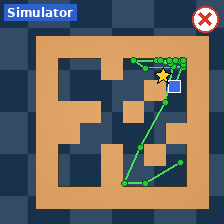}
    \caption{Low Mass \\ frozen $\rightarrow$ failure}
  \end{subfigure}
  \begin{subfigure}{0.16\textwidth}
    \includegraphics[width=\linewidth]{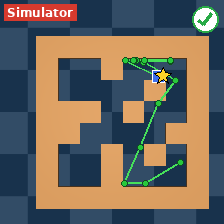}
    \caption{Low Mass \\ adapt $\rightarrow$ success}
  \end{subfigure}\hfill
  \begin{subfigure}{0.16\textwidth}
    \includegraphics[width=\linewidth]{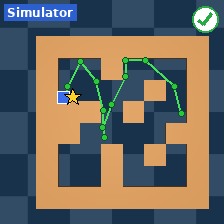}
    \caption{Default Dyn. \\ frozen $\rightarrow$ success}
  \end{subfigure}
  \begin{subfigure}{0.16\textwidth}
    \includegraphics[width=\linewidth]{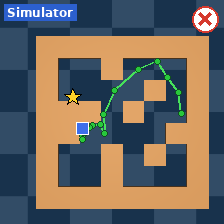}
    \caption{High Damping \\ frozen $\rightarrow$ failure}
  \end{subfigure}
  \begin{subfigure}{0.16\textwidth}
    \includegraphics[width=\linewidth]{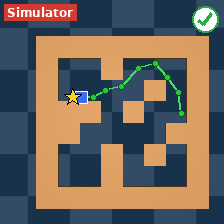}
    \caption{High Damping \\ adapt $\rightarrow$ success}
  \end{subfigure}
  \captionsetup{type=figure}
  \caption{PointMaze-Medium Dynamics-Shift Planning Trajectories. The green polylines trace the agent's position over time, the blue square marks the end, and the gold star marks the goal. Under in-distribution dynamics the model reaches the goal. However, frozen world model mispredicts and planning fails under dynamics shifts, while test-time adaptation realigns with the new environment and recovers success.}
  \label{fig:medium_dynamics_ood_examples}

  \setcounter{subfigure}{0}%
  \renewcommand{\thesubfigure}{\ifcase\value{subfigure}\or a1\or a2\or b1\or b2\or c1\or c2\fi}%
  \begin{subfigure}{0.16\textwidth}
    \includegraphics[width=\linewidth]{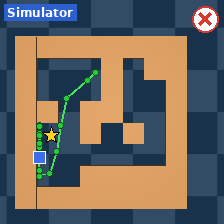}
    \caption{Maze1 \\ frozen $\rightarrow$ failure}
  \end{subfigure}
  \begin{subfigure}{0.16\textwidth}
    \includegraphics[width=\linewidth]{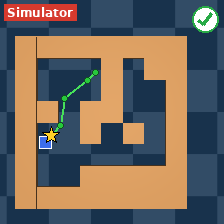}
    \caption{Maze1 \\ adapt $\rightarrow$ success}
  \end{subfigure}\hfill
  \begin{subfigure}{0.16\textwidth}
    \includegraphics[width=\linewidth]{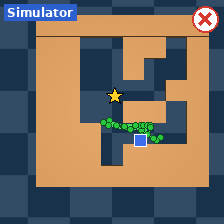}
    \caption{Maze2 \\ frozen $\rightarrow$ failure}
  \end{subfigure}
  \begin{subfigure}{0.16\textwidth}
    \includegraphics[width=\linewidth]{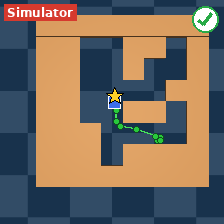}
    \caption{Maze2 \\ adapt $\rightarrow$ success}
  \end{subfigure}\hfill
  \begin{subfigure}{0.16\textwidth}
    \includegraphics[width=\linewidth]{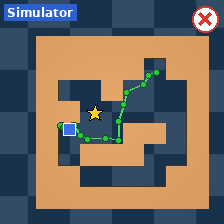}
    \caption{Maze3 \\ frozen $\rightarrow$ failure}
  \end{subfigure}
  \begin{subfigure}{0.16\textwidth}
    \includegraphics[width=\linewidth]{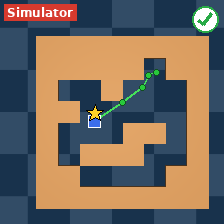}
    \caption{Maze3 \\ adapt $\rightarrow$ success}
  \end{subfigure}
  \captionsetup{type=figure}
  \caption{Diverse Maze Planning Trajectories. We use the same visual conventions as in \Cref{fig:medium_dynamics_ood_examples}. While the frozen model fails on unseen mazes, AdaJEPA succeeds with trajectories close to the shortest paths.}
  \label{fig:diverse_maze_examples}
\vspace{-0.3in}
\end{figure*}

\paragraph{In-distribution performance.}
We first examine how AdaJEPA behaves on test environments same as training. On PushObj training shapes and the default PushT rendering, test-time adaptation significantly improves over the frozen model for both GD and CEM planners as shown in~\Cref{fig:pushobj}. As PushObj model is trained on various shapes, adaptation to the current shape makes the model specialize and results in the largest performance boost of over 20\% gain. On PointMaze with default dynamics, adaptation preserves the strong frozen-model baseline~\Cref{tab:pointmaze_layout_dynamics}. In summary, test-time adaptation is
safe to apply in-distribution: it yields large gains when the frozen model is suboptimal and does no harm when it is already near-optimal.

\paragraph{Distribution shifts.}
AdaJEPA gives consistent and significant gains on out-of-distribution test environments across all settings. Its success continues to increase over replanning steps, whereas the frozen model often saturates early, showing that test-time adaptation helps the planner recover from initially inaccurate predictions (\Cref{fig:pushobj}). On unseen shapes of PushObj, the frozen model's performance drops substantially, while AdaJEPA nearly doubles the planning success rate. For visual shifts, AdaJEPA improves robustness to common visual changes with clear gains under blur, noise, lighting. However, gains are modest under the red-anchor and red-block shifts, likely because the model relies on color to distinguish the fixed anchor from the manipulated object, which may requires data augmentation or explicit invariance regularization. On dynamics shifts in~\Cref{tab:pointmaze_layout_dynamics}, it is interesting that the frozen model already performs strongly on new dynamics, which might be due to in context learning on the history of three frames. AdaJEPA still shows consistent additional gains beyond the strong baseline, For unseen mazes in layout shifts, the default \texttt{predlast + enclast} update improves over the frozen model, and adapting earlier predictor layers improves further. Furthermore, for both maze environments, the resulting planning trajectories are also closer to the shortest path after adaptation (\Cref{fig:medium_dynamics_ood_examples,fig:diverse_maze_examples}). Overall, these results show that lightweight test-time adaptation is very effective in improving world models under distribution shift.

\paragraph{Latency.}
While the adaptation step significantly increases the planning success rate, it inevitably increases latency due to extra parameter update. However, the increased time is almost negligible (\Cref{tab:pusht_ttt}) as we only update a small subset of parameters using only one step. Moreover, adaptation also makes the agent reach the goal in equal-or-fewer replans, thus decreasing the overall time. These results show that AdaJEPA's lightweight test-time adaptation is both efficient and effective.

\begin{figure*}[t]
\centering
\noindent
\makebox[\linewidth][c]{%
\begin{minipage}[c]{0.04\linewidth}
    \makebox[\linewidth][r]{\rotatebox[origin=c]{90}{\footnotesize (a) PushT-RedAnchor}}
\end{minipage}%
\hspace{3pt}%
\begin{minipage}[c]{0.96\linewidth}
    \includegraphics[width=\linewidth]{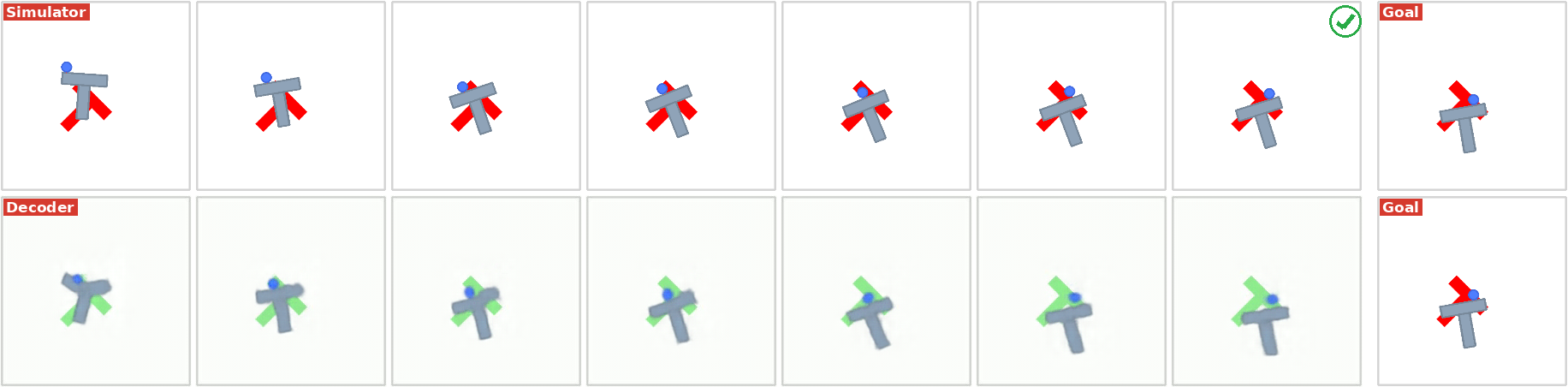}
    \label{fig:exp_red}
\end{minipage}%
}

\vspace{1mm}

\makebox[\linewidth][c]{%
\begin{minipage}[c]{0.04\linewidth}
    \makebox[\linewidth][r]{\rotatebox[origin=c]{90}{\footnotesize (b) PushObj-Square}}
\end{minipage}%
\hspace{3pt}%
\begin{minipage}[c]{0.96\linewidth}
    \includegraphics[width=\linewidth]{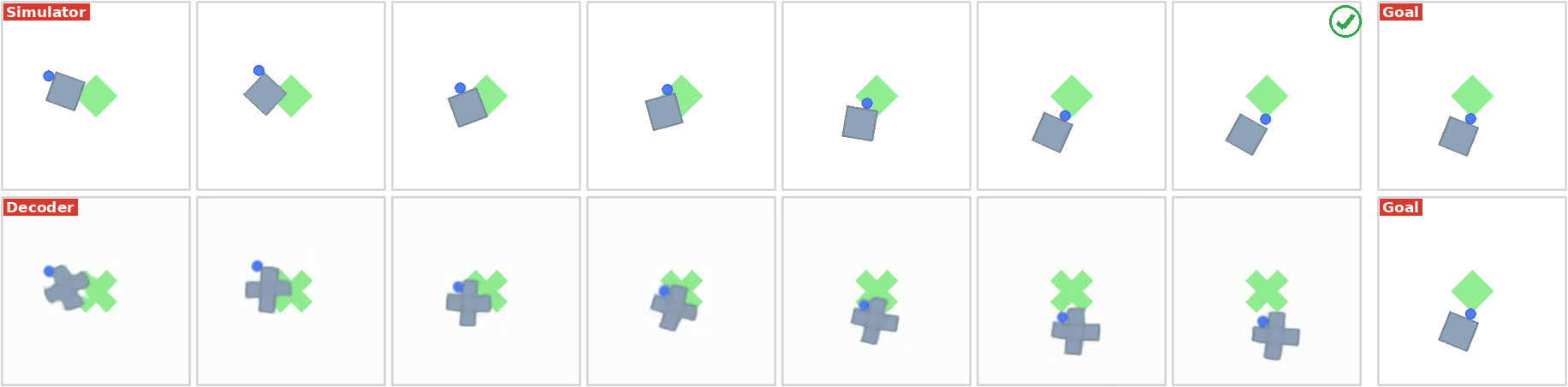}
\end{minipage}%
}
\caption{Examples of AdaJEPA Planning Trajectories under Visual Shifts and Shape Shifts. The decoder is trained on pretrained representations; i.e. (a) original PushT data with a gray pushed block; (b) PushObj data which only includes \{\texttt{T}, \texttt{L}, \texttt{Z}, \texttt{+}\} while \texttt{square} is an unseen shape. }
\label{fig:exp}
\end{figure*}

\paragraph{Visualization. }
To better understand what test-time adaptation changes in the world model, we visualize imagined rollouts after adaptation. We train a decoder on the pretrained latent space and find it can still reconstruct visual rollouts after light-weight test-time adaptation. Under visual corruptions or unseen PushObj shapes, decoded rollouts often retain training-domain structure: for example, an unseen red PushT block may be decoded as a gray block (the training color), and an unseen object may be decoded as a visually similar seen shape~(\Cref{fig:exp}, with more examples in~\Cref{app:viz}). This suggests that AdaJEPA improves planning by exploiting shared latent structure and recalibrating predictions, while remaining close to the learned latent manifold.

\begin{table*}[t]
\centering
\caption{AdaJEPA across different implementations. The planning success is evaluated using PushT validation trajectories from~\citet{dinowm}. The reported values are the average MPC success rate (\%) and per-replan time using one H200. Test-time adaptation consistently improves and introduces almost negligible latency. }
\label{tab:pusht_ttt}
\resizebox{\linewidth}{!}{%
\begin{tabular}{l c c l
                r@{\,{\tiny$\pm$}\,}l r@{\;}l
                r@{\,{\tiny$\pm$}\,}l r@{\;}l}
\toprule
Encoder / Predictor & Latent Dim & Anti-collapse & Setting
& \multicolumn{2}{c}{GD (\%)} & \multicolumn{2}{c}{Time (s)}
& \multicolumn{2}{c}{CEM (\%)} & \multicolumn{2}{c}{Time (s)} \\
\midrule

WM w/ Temporal Straightening
& \multirow{2}{*}{$1{\times}384$}
& \multirow{2}{*}{stop-grad}
& Frozen
& 84.0 & {\tiny 2.0}
& 3.14 &
& 74.0 & {\tiny 3.5}
& 0.24 & \\
(global feat.) \citep{wang2026temporal_straightening}
& &
& Adapt
& 85.3 & {\tiny 3.1}\,{\scriptsize\textcolor{LabBlue}{$\uparrow$1.3}}
& 3.17 & {\scriptsize\textcolor{red}{$\uparrow$0.03}}
& 81.3 & {\tiny 6.4}\,{\scriptsize\textcolor{LabBlue}{$\uparrow$7.3}}
& 0.27 & {\scriptsize\textcolor{red}{$\uparrow$0.03}} \\
\midrule

WM w/ Temporal Straightening
& \multirow{2}{*}{$196{\times}384$}
& \multirow{2}{*}{stop-grad}
& Frozen
& 91.3 & {\tiny 4.2}
& 3.37 &
& 89.3 & {\tiny 3.1}
& 5.37 & \\
(spatial feat.) \citep{wang2026temporal_straightening}
& &
& Adapt
& 92.0 & {\tiny 3.5}\,{\scriptsize\textcolor{LabBlue}{$\uparrow$0.7}}
& 3.38 & {\scriptsize\textcolor{red}{$\uparrow$0.01}}
& 93.3 & {\tiny 2.3}\,{\scriptsize\textcolor{LabBlue}{$\uparrow$4.0}}
& 5.39 & {\scriptsize\textcolor{red}{$\uparrow$0.02}} \\
\midrule

DINO-WM (patch)
& \multirow{2}{*}{$196{\times}384$}
& \multirow{2}{*}{-}
& Frozen
& 68.0 & {\tiny 10.6}
& 3.66 &
& 86.7 & {\tiny 6.1}
& 9.53 & \\
(spatial feat.) \citep{dinowm}
& &
& Adapt
& 70.0 & {\tiny 4.0}\,{\scriptsize\textcolor{LabBlue}{$\uparrow$2.0}}
& 3.68 & {\scriptsize\textcolor{red}{$\uparrow$0.02}}
& 90.0 & {\tiny 3.5}\,{\scriptsize\textcolor{LabBlue}{$\uparrow$3.3}}
& 9.56 & {\scriptsize\textcolor{red}{$\uparrow$0.03}} \\
\bottomrule
\end{tabular}%
}
\end{table*}

\subsection{Ablations}
\label{ablations}

\paragraph{Which parameters to adapt?}

We compare direct updates to selected predictor and encoder layers and LoRA updates to the full model (full results are in~\Cref{fig:layer_ablation}). Across shifts, all adaptation choices improve over the frozen model, indicating that the superior performance of AdaJEPA is not tied to a specific adaptation target. For shape shifts, all variants perform similarly, even when the encoder is frozen, suggesting that the pretrained representation shows generalization across object geometries and that most of the needed correction lies in the predictor. For visual and layout shifts, updating only the predictor is less effective, as the mismatch enters through the observation representation and cannot be fully corrected with predictor alone. Interestingly, adapting the first layer of the predictor is particularly effective to Layout Shifts, likely because this layer is closest to the latent and action inputs and can better recalibrate local transition structure under new maze connectivity. LoRA also improves over the frozen model, but does not consistently outperform direct updates to selected layers. Overall, the best adaptation target is \emph{environment-dependent, but performance is not highly sensitive}: simple selected-layer updates, such as adapting the first or last predictor block together with the last encoder stage, are already strong across settings.

\paragraph{How should adaptation hyperparameters be chosen?}
Since AdaJEPA adapts inside the MPC loop, the update must be effective with little tuning and minimal latency. We ablate the test-time learning rate, number of gradient steps, and replay buffer in~\Cref{fig:abalations_lr_buffer}. 
\emph{A single gradient step with the training learning rate provides a strong practical default}: larger learning rates can make one-step updates effective but become less stable with more steps, while smaller learning rates often require additional updates and increase replanning cost. Replay buffer design has a smaller effect: AdaJEPA improves over the frozen model across buffer choices, including no buffer, with a recent-transition buffer giving the most stable gains. For out-of-distribution settings, stronger adaptation---via a larger learning rate, more optimization steps, or a larger recent replay buffer---is often beneficial when latency permits. These hyperparameters can be tuned for a target environment if necessary, while the default provides a practical starting point.

\paragraph{AdaJEPA improves over various JEPA world models.} 
We evaluate AdaJEPA across multiple JEPA world-model implementations on the PushT validation trajectories (details in~\Cref{app:pusht}). As shown in~\Cref{tab:pusht_ttt}, \emph{AdaJEPA improves planning success across global and spatial representations, model architectures, GD and CEM planners, and different training objectives}. Even though these models are well trained and evaluated in distribution, test-time adaptation still yields consistent gains while adding only \(0.01\)--\(0.03\)s per MPC replanning step. These results indicate that AdaJEPA is broadly efficient and effective for improving the test-time performance for latent world models.

\subsection{Discussion}
\label{sec:discussion}

The experiments above show that AdaJEPA consistently improves planning by correcting prediction errors during closed-loop MPC, and is robust across environments, planners, base models, and adaptation targets.

\begin{wrapfigure}{r}{0.6\linewidth}
    \centering
    \vspace{-2em}
    \includegraphics[width=\linewidth]{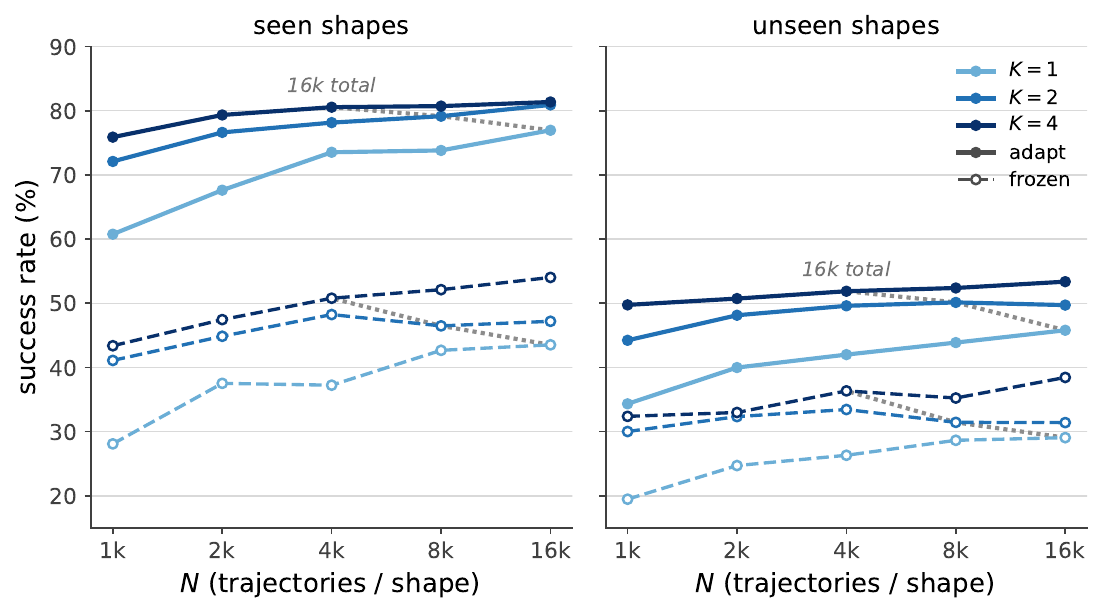}
    \vspace{-2em}
    \caption{
    Effect of training data scale on PushObj planning success: shape diversity $K$ and trajectories per shape $N$.
    }
    \vspace{-2em}
    \label{fig:scale}
\end{wrapfigure}

To further understand the strength and limitations of AdaJEPA, we study how training data scale affects generalization and adaptivity. We use PushObj as a test bed, and analyze two axes: shape diversity $K$ and trajectories per shape $N$. We report the average success on seen and unseen shapes across different scales in~\Cref{fig:scale}, with more experiment details in~\Cref{app:scale}. In general, larger and more diverse training data improves both frozen and AdaJEPA across seen and unseen shapes. Shape diversity is especially important for generalization before and after adaptation: for example, under the same $16$k total trajectory budget, distributing data across four shapes $(K{=}4,N{=}4\text{k})$ achieves $51.9\%$ unseen success with AdaJEPA, compared with $45.8\%$ when all trajectories come from a single shape $(K{=}1,N{=}16\text{k})$. Test-time adaptation improves success rates across scales, especially in \emph{low-data} regimes: on seen shapes with $K{=}1,N{=}1\text{k}$, it improves success from $28.1\%$ to $60.8\%$, more than doubling performance over the frozen model and surpassing a frozen model trained with $16{\times}$ more trajectories per shape ($43.5\%$). Thus, while more training data improves generalization, \emph{test-time adaptation can compensate for limited training coverage} by refining the model during deployment, providing a more sample-efficient route that is complementary to data scaling alone. 

However, as we only perform lightweight correction during planning, its effectiveness is also bounded by the coverage of the pretrained representation: when the test environment requires features absent from training, adaptation can improve planning but may not fully close the gap. A natural next step is to combine lightweight test-time adaptation with continual and active learning to expand the world model's coverage over time.

\section{Conclusion}

We introduced AdaJEPA, an adaptive world model that performs test-time adaptation during the closed-loop MPC. This creates a simple plan--execute--adapt--replan loop: the model uses predictions to select actions, the actions lead to new observations from the environment, the new observations are used to update the model, and the updated model is immediately reused for the next planning step with better predictions. Our experiments show that a single gradient step per transition is sufficient to recover substantial planning performance under both visual and dynamics shifts. More broadly, latent world models should continue to be trained at deployment, rather than kept frozen. We believe this work opens a promising direction for adaptive world models that continuously calibrate predictions and update representations while acting, enabling more resilient perception and planning in a changing world.
\section*{Acknowledgments}
We thank Gaoyue Zhou and Daohan Lu for helpful discussions. This work was supported in part by AFOSR under grant FA95502310139, NSF Awards 1922658 and 2545541, Visko Platform, a Google TPU Award, Toyota Research Institute R2I program, the NYU-KAIST Award A25-0081-002, and the Institute of Information \& Communications Technology Planning Evaluation (IITP) under grant RS-2024-00469482, funded by the Ministry of Science and ICT (MSIT) of the Republic of Korea in connection with the Global AI Frontier Lab International Collaborative Research.
The compute is supported by the NYU High Performance Computing resources, services, and staff expertise.

\bibliographystyle{apalike}
\bibliography{main}
\newpage

\crefname{appendix}{Appendix}{Appendices}
\Crefname{appendix}{Appendix}{Appendices}
\crefalias{section}{appendix}
\crefalias{subsection}{appendix}
\appendix

\section{Environments and Data}
\label{app:env}

\subsection{PushT (DINO-WM)}
\label{app:pusht}

PushT is a contact-rich manipulation environment introduced by~\citep{chi2024pusht}. It consists of a circular pusher agent interacting with a T-shaped block. Starting from a random initial state, the agent must push the block to a target pose. Here, we use the setup from DINO-WM~\citep{dinowm}, which treats the fixed green T as a visual reference rather than the target object itself. We use DINO-WM's PushT validation trajectories to compare frozen JEPA world models and their adaptive counterparts in~\Cref{tab:pusht_ttt}.

\subsection{PushObj}
\label{app:pushobj}

PushObj extends PushT by replacing the T-shaped block with objects of different shapes, \{\texttt{L}, \texttt{Z}, \texttt{+}, \texttt{I}, \texttt{smallT}, \texttt{square}\}, following~\citet{dinowm}. However, simply replacing the object while replaying the original action sequences yields trajectories with much sparser contact: actions collected for the T-shaped block often no longer align with the new geometry, causing the agent to miss the object.

To construct PushObj, we start from \(N=18{,}500\) PushT training trajectories and introduce a contact bias to increase the fraction of trajectories in which the pusher interacts with the object. We generate \(16{,}000\) training trajectories per shape. For testing, we filter the remaining trajectories to keep only those containing at least one contact. This makes the evaluation meaningful by filtering out trivial no-contact motion. We denote this dataset as \texttt{PushObj-all-16k} and have the following subsets for major experiments:
\begin{itemize}
    \item \texttt{PushObj-TLZ+-4k}: we train on four shapes \{\texttt{T}, \texttt{L}, \texttt{Z}, \texttt{+}\} (each with 4k trajectories), and evaluate on the test trajectories of all seven shapes. This data is used in the \textbf{Shape Shifts} experiments shown in the upper row of~\Cref{fig:pushobj}. We train the model for 3 epochs.
    \item \texttt{PushObj-T-16k}: we train on shape T (16k trajectories) only, and evaluate on various visual corruptions of its test trajectories. This data is used in the \textbf{Visual Shifts} experiments shown in the lower row of~\Cref{fig:pushobj}. We train for 3 epochs.
\end{itemize}

\subsection{PointMaze-Medium}
\label{app:pointmaze}

PointMaze is a 2D navigation environment based on the MuJoCo physics engine~\citep{fu2020d4rl}. The task is to navigate from a start position to a target position, specified by start and goal images. The action space consists of forces applied along the $x$ and $y$ axes. We use Medium-Maze data from~\citet{wang2026temporal_straightening} and directly use their pretrained checkpoint of ResNet global features for the \textbf{Dynamics Shifts} experiments as shown in~\Cref{tab:pointmaze_layout_dynamics}. Unlike prior works that simply sample goal positions that are $T$ steps away from the start in the test trajectories, we impose nontrivial goals in this environment. For PointMaze-Medium, there are 26 open grid cells of the medium maze. We start by picking two distinct cells without replacement, and then reject and resample until their Euclidean distance is larger than 3 cell units.

\subsection{Diverse PointMaze}
\label{app:diversemaze}

We largely follow the data-generation procedure of HWM-PLDM~\citep{zhang2026hierarchical} to generate 30 PointMaze environments with random \(8 \times 8\) maze layouts where 25 are used for training (2,000 trajectories each, 50,000 in total) and 5 held-out layouts are saved for evaluation. At test time, we evaluate on the held-out layouts with 50 episodes in total, 10 per layout. For each episode, the start cell is sampled uniformly from the open cells, and the goal cell is sampled at a controlled shortest-path distance of 3--5 cells (computed by BFS). This is used for the \textbf{Layout Shifts} experiments in~\Cref{tab:pointmaze_layout_dynamics}. We train for 3 epochs.

\vspace{-0.1in}

\section{Experiments}
\label{app:exp}

\subsection{Hyperparameters}

\begin{table}[h]
\centering
\begin{minipage}[t]{0.45\textwidth}
\centering
\caption{Training Hyperparameters. }
\label{tab:train_hyperparams}
\begin{tabular}{ll}
\hline
\textbf{Name} & \textbf{Value} \\ 
\hline
Encoder lr & 1e-5 \\
Predictor lr & 5e-4 \\
Action/Prop encoder lr & 5e-4 \\
Batch size & 64 \\ 
History frames & 3 \\
Frameskip & 5 \\
\hline
\end{tabular}
\end{minipage}
\hfill
\begin{minipage}[t]{0.45\textwidth}
\centering
\caption{Planning Hyperparameters.}
\label{tab:plan_hyperparams}
\begin{tabular}{ll}
\hline
\textbf{Name} & \textbf{Value} \\ 
\hline
Subplanner horizon & 25 \\
\# Executed actions & 5 \\
\hline
GD Optimizer & Adam \\
GD Action Initialization & Zero \\
GD Learning rate & 0.1 \\
GD \#opt steps & 100 \\
\hline
CEM \# samples & 200 \\
CEM \#opt steps & 10 \\
\hline
\end{tabular}
\end{minipage}
\end{table}

\vspace{-0.3in}
\subsection{Ablations}
\label{app:ablations}

\begin{figure*}[t]
  \centering
  \includegraphics[width=\textwidth]{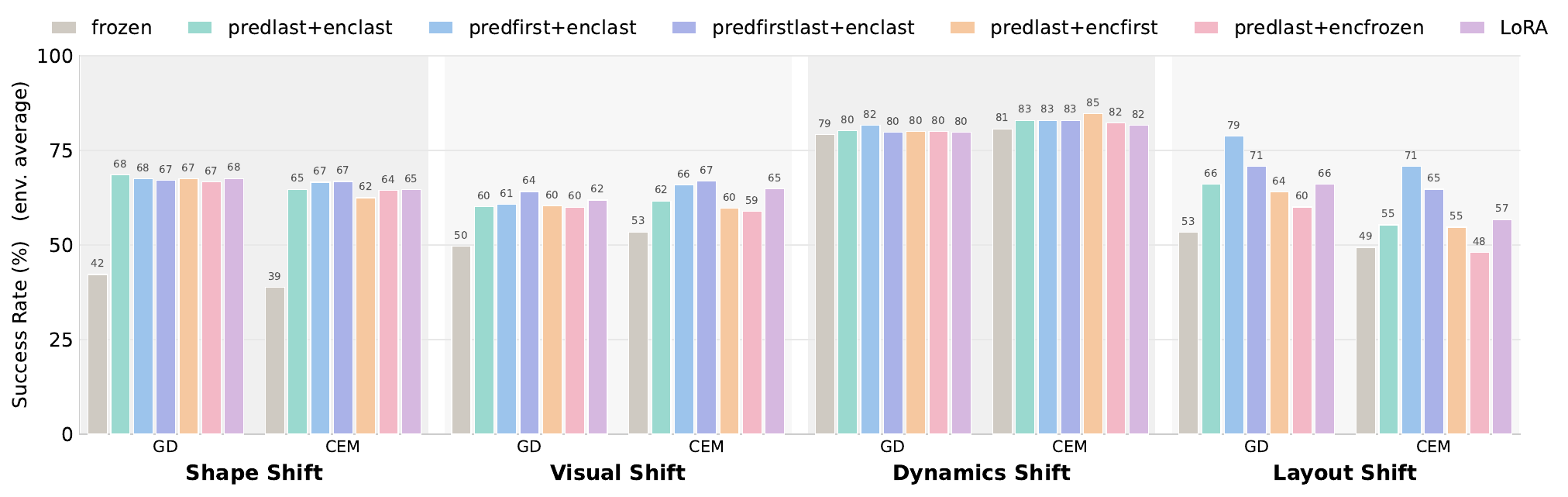}
  \caption{
    Effect of Adaptation Layers to Planning Success Rates. The reported values are the per-shift success rates (\%) averaged over all setups within each shift. Test-time adaptation improves planning across all distribution shifts and is largely insensitive to which layers are adapted.}
  \label{fig:layer_ablation}
\end{figure*}

\begin{figure*}[t]
    \centering
    \begin{subfigure}{0.48\linewidth}
        \centering
        \includegraphics[width=\linewidth]{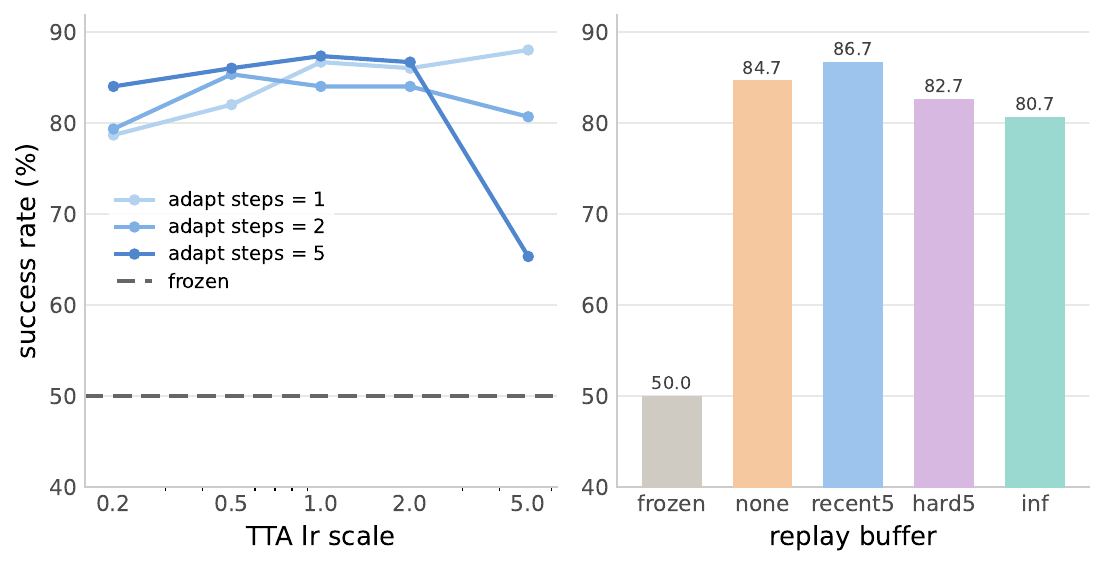}
        \caption{T (seen shape)}
        \label{fig:ablations_t_lr}
    \end{subfigure}
    \hfill
    \begin{subfigure}{0.48\linewidth}
        \centering
        \includegraphics[width=\linewidth]{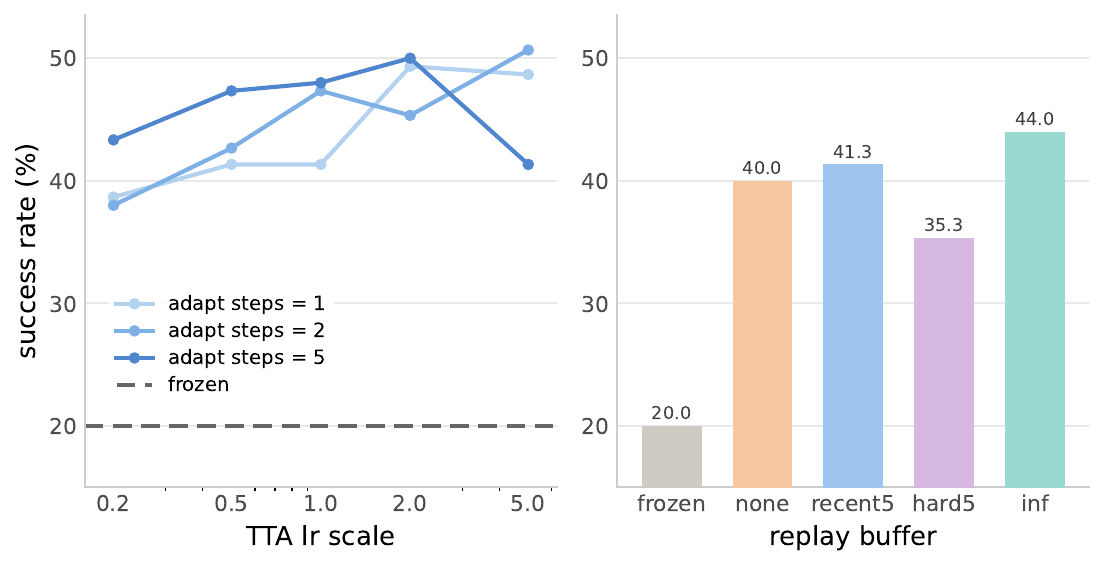}
        \caption{Square (unseen shape)}
        \label{fig:ablations_square_lr}
    \end{subfigure}
    \caption{Effect of Adaptation Hyperparameters and Replay Buffers to Planning Success Rates.}
    \label{fig:abalations_lr_buffer}
\end{figure*}

We ablate the main design choices in AdaJEPA's test-time adaptation, including which parameters are updated, the adaptation learning rate, the number of gradient steps, and the replay-buffer design.

\paragraph{Setup.} The predictor is a ViT-style module, consisting of a stack of transformer blocks followed by a final LayerNorm. The encoder is a small ResNet whose stages, in order, are five residual blocks (\texttt{rb1}--\texttt{rb5}), an optional pooling head, and a projection head. We define the encoder's \emph{first stage} as \texttt{rb1}, the input residual block mapping \(3\!\rightarrow\!32\) channels and extracting low-level features. We define its \emph{last stage} as the projection head, a Linear--GELU--Linear--LayerNorm module that maps pooled features to the latent embedding. Unless otherwise specified, every adapting variant uses the same protocol: one gradient step per MPC replanning step, a replay buffer containing the five most recent transitions, a predictor learning rate of \(5\times10^{-4}\), and an encoder learning rate of \(10^{-5}\). We use a maximum MPC steps of 20 (instead of 30 in the main experiments) to speed up experiments.

\paragraph{What to adapt.} In general, AdaJEPA is robust to the choice of adaptation target. As shown in~\Cref{fig:layer_ablation}, all adaptation variants improve over the frozen model across shape, visual, dynamics, and layout shifts. Updating only a small subset of parameters is often sufficient: \texttt{predlast+enclast} is consistently competitive, while adapting earlier predictor layers can further help under layout shifts. LoRA also improves over the frozen model but does not consistently outperform direct adaptation of selected predictor and encoder layers. The different variants are described below.

\begin{itemize}
  \item \texttt{Frozen.} No test-time adaptation. The encoder and predictor are kept fixed throughout planning.

  \item \texttt{predlast+enclast}. Adapts the predictor's \emph{last} transformer block, including the final LayerNorm, and the encoder's \emph{last} stage.
  \item \texttt{predfirst+enclast}. Adapts the predictor's \emph{first} transformer block and the encoder's \emph{last} stage.
  \item \texttt{predfirstlast+enclast}. Adapts the predictor's \emph{first} and \emph{last} transformer blocks, including the final LayerNorm, and the encoder's \emph{last} stage.
  \item \texttt{predlast+encfirst}. Adapts the predictor's \emph{last} transformer block, including the final LayerNorm, and the encoder's \emph{first} stage.
  \item \texttt{predlast+encfrozen}. Adapts only the predictor's \emph{last} transformer block, including the final LayerNorm. The encoder remains frozen.
  \item \texttt{LoRA.} Inserts low-rank adapters with rank \(8\) and \(\alpha=16\) into every linear layer of the full predictor and encoder. Only the adapter parameters are updated, while all base weights remain frozen.
\end{itemize}

\paragraph{Learning rates, optimization steps, and replay buffers.}
We ablate test-time adaptation hyperparameters and replay-buffer design on PushObj using a seen shape, \texttt{T}, and a held-out shape, \texttt{square} (\Cref{fig:abalations_lr_buffer}). Across the grid, adaptation is consistently effective: every setting substantially outperforms the frozen world model, improving success from \(50\%\) to \(88\%\) on \texttt{T}, and from \(20\%\) to \(51\%\) on \texttt{square}. 

The learning rate and the number of gradient steps are tightly coupled. A large learning rate (\(5\times\) the training rate) is highly effective with a single step, but can overshoot when combined with multiple steps. Conversely, a small learning rate (\(0.2\times\)) is more stable but requires more updates to become competitive, increasing latency. Moderate settings (\(1\)--\(2\times\), one or two steps) are robust across both shapes, although the optimum remains environment- and model-dependent. To keep adaptation simple and efficient, we use the training learning rate and take one gradient step per MPC replanning step by default, which is competitive on both shapes at minimal cost.

Replay-buffer design has a smaller effect: success varies only moderately across buffer choices, ranging from \(81\%\)--\(87\%\) on \texttt{T} and \(35\%\)--\(44\%\) on \texttt{square}. Importantly, every replay-buffer variant, including no buffer, substantially outperforms the frozen model. We use a recent sliding-window buffer in the main experiments, as it provides the most stable gains.

\subsection{Additional Experiments: Training Data Scale for PushObj}
\label{app:scale}

\begin{figure}[t]

    \centering
    \includegraphics[width=0.6\linewidth]{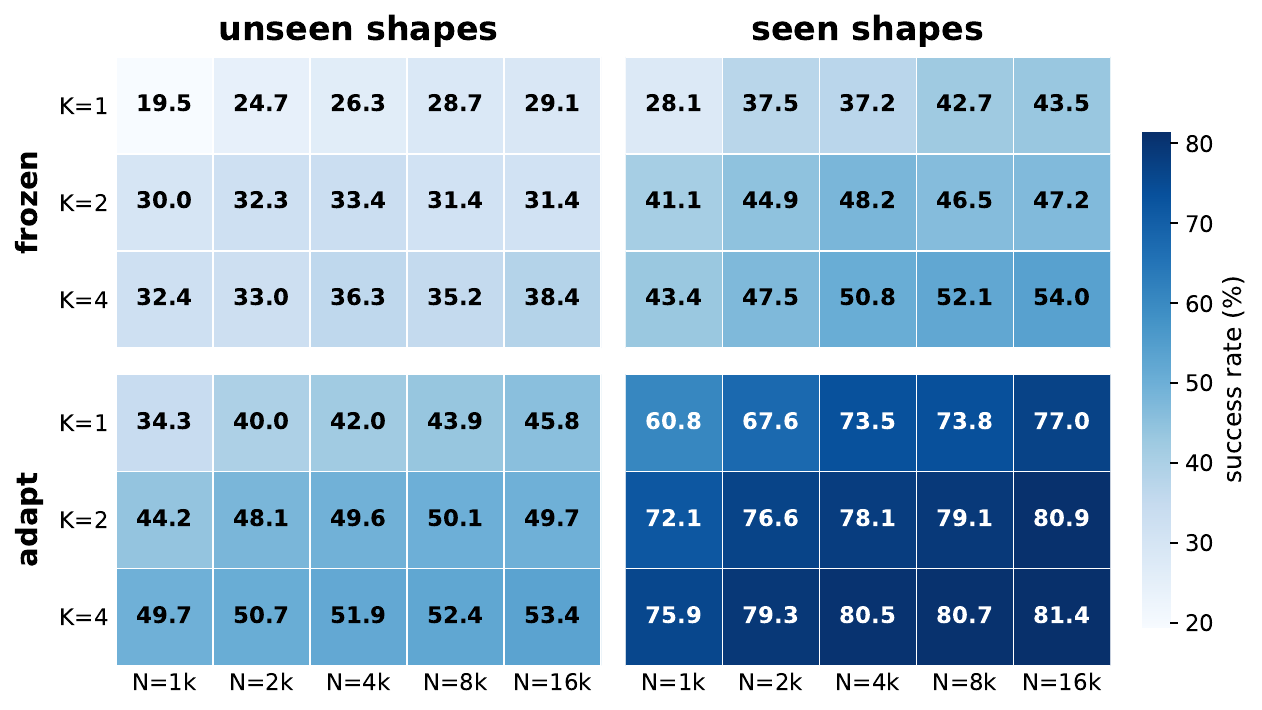}
    \caption{
    Heatmap of offline training data scale vs. PushObj planning success.
    Rows vary shape diversity $K$ and columns vary trajectories per shape $N$.
    }
    \label{fig:scale_heatmap}
\end{figure}

We study how training-data size and diversity affect test-time adaptation on PushObj, which contains seven object shapes. We vary the number of training shapes, \(K\in\{1,2,4\}\), and the number of trajectories per shape, \(N\in\{1\text{k},2\text{k},4\text{k},8\text{k},16\text{k}\}\). For each \(K\), we construct seven balanced training sets, each containing \(K\) shapes, so that every shape appears in exactly \(K\) sets. We train one world model per set using \(K\times N\) trajectories, keeping the number of optimization steps fixed across settings to isolate the effect of data. At test time, we evaluate each model on all seven shapes, label each shape as \emph{seen} or \emph{unseen} according to the corresponding training set, and report success averaged over three seeds and over shapes in each group. We plot the results using line charts in~\Cref{fig:scale} and heatmaps in~\Cref{fig:scale_heatmap}.

\begin{itemize}
    \item \textbf{Both more trajectories and more shape diversity are beneficial}: the frozen model improves with both more trajectories and more shape diversity: from \((K{=}1,N{=}1\text{k})\) to \((K{=}4,N{=}16\text{k})\), success increases from \(28\%\) to \(54\%\) on seen shapes and from \(20\%\) to \(38\%\) on unseen shapes.
    \item \textbf{Diversity matters more than the number of trajectories per shape}: with $16$k trajectories total, distributing them over four shapes ($K{=}4,N{=}4\text{k}$) outperforms concentrating them on a single shape ($K{=}1,N{=}16\text{k}$) on both seen ($81\%$ vs.\ $77\%$ adapted, $51\%$ vs.\ $44\%$ frozen) and unseen shapes ($52\%$ vs.\ $46\%$ adapted, $36\%$ vs.\ $29\%$ frozen).
    \item \textbf{AdaJEPA provides a consistent gain at every data scale and can even compensate for large reductions in training data}: test-time adaptation leads to an average improvement of over 30\% on seen shapes and over 15\% on unseen shapes. Notably, adaptation can offset large reductions in training data. On seen shapes, the adapted model trained on a single shape with only \(1\)k trajectories reaches \(61\%\), exceeding the largest frozen model trained on four shapes with \(16\)k trajectories each (\(64\)k total, \(54\%\)). On unseen shapes, an adapted single-shape model already surpasses the best four-shape frozen model once \(N\ge2\)k. 
\end{itemize}

In general, test-time adaptation is most valuable when data is scarce, yet its benefits persist as data scales, providing a sample-efficient path to robust generalization.

\section{Visualization of Planning Trajectories}
\label{app:viz}

Here, we provide a qualitative comparison of single planning episodes, comparing the frozen model and AdaJEPA under identical start/goal conditions. 

For each example, the upper rows show the trajectory executed for each method. The \emph{simulator} row shows the rendered environment, while the \emph{decoder} row shows the model's imagined latent rollout decoded to pixels. We show the frozen run for 13 steps (capped for better visibility), and gray out AdaJEPA steps after it reaches the goal. A check mark or cross indicates planning success or failure. The bottom panel reports the latent prediction loss at each replanning step. Before the current adaptation update, the model predicts a short latent rollout; we compare these predictions to the latents encoded from the observations actually reached by executing the planned actions. AdaJEPA reduces this loss by continuously refining the model's predictions, whereas the frozen model often fails due to inaccurate predictions.

\begin{figure}[h]
    \centering
    \includegraphics[width=\textwidth]{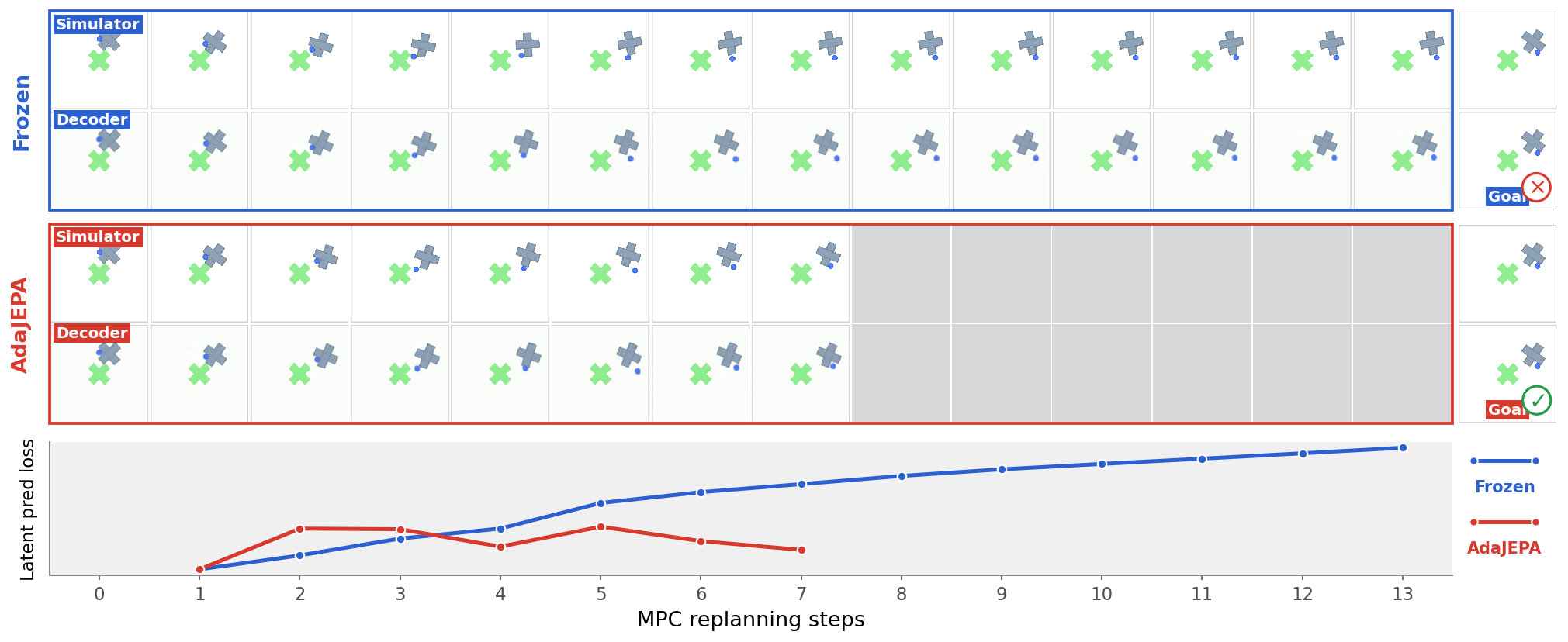}
    \vspace{-0.3in}
    \caption{The model is trained on shape \{T,L,+,Z\}, and tested on a seen shape + here.}
    \vspace{-0.1in}
    \label{fig:example}
\end{figure}

\begin{figure}[h]
    \centering
    \includegraphics[width=\textwidth]{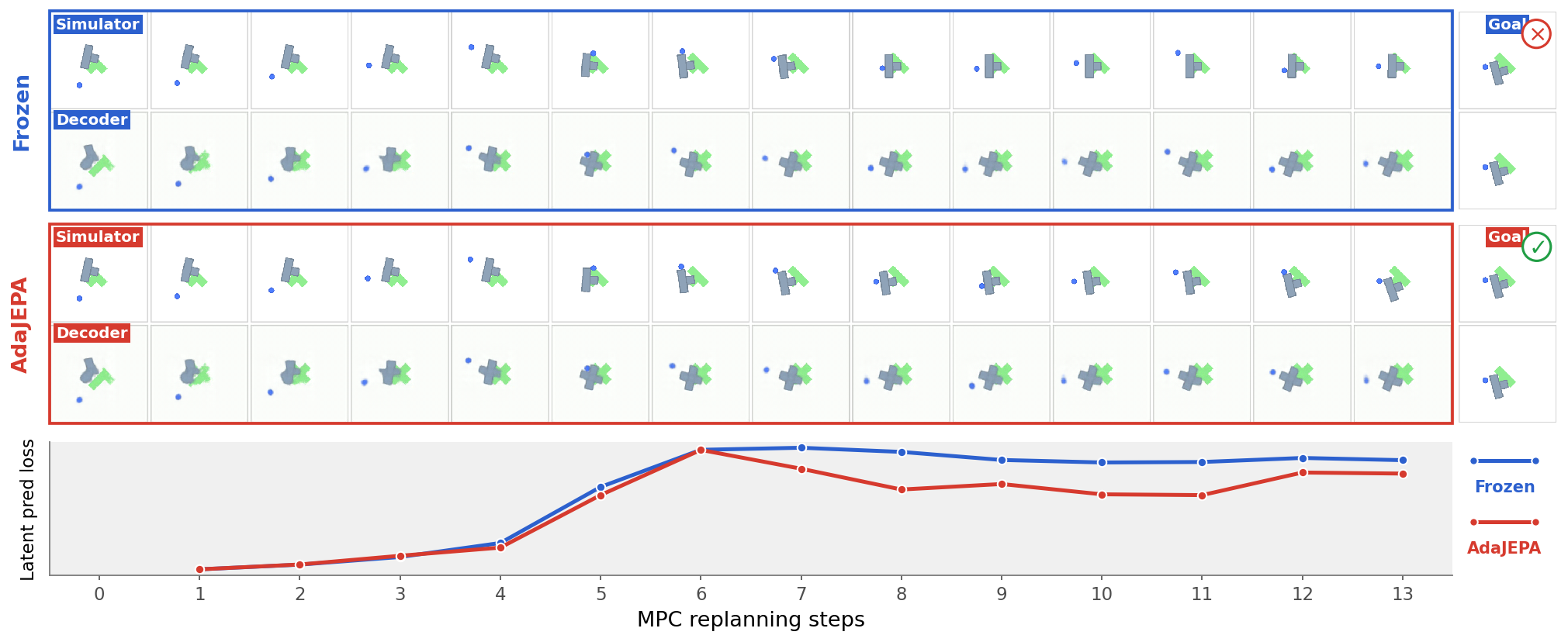}
    \vspace{-0.3in}
    \caption{Shape Shifts: The model is trained on shapes \{\texttt{T}, \texttt{L}, \texttt{Z}, \texttt{+}\}, and tested on an unseen shape \texttt{smallT}. The decoder tries to decode to the closest seen shapes.}
    \vspace{-0.2in}
    \label{fig:example}
\end{figure}

\newpage

When facing visual corruptions and color changes at test time, the representation and prediction also need adaptation. AdaJEPA reduces the prediction loss consistently and leads to better planning. Since the decoder is only trained on default PushT data, it tends to reconstruct the original colors and scenes.

\vspace{0.1in}

\begin{figure}[h]
    \centering
    \includegraphics[width=\textwidth]{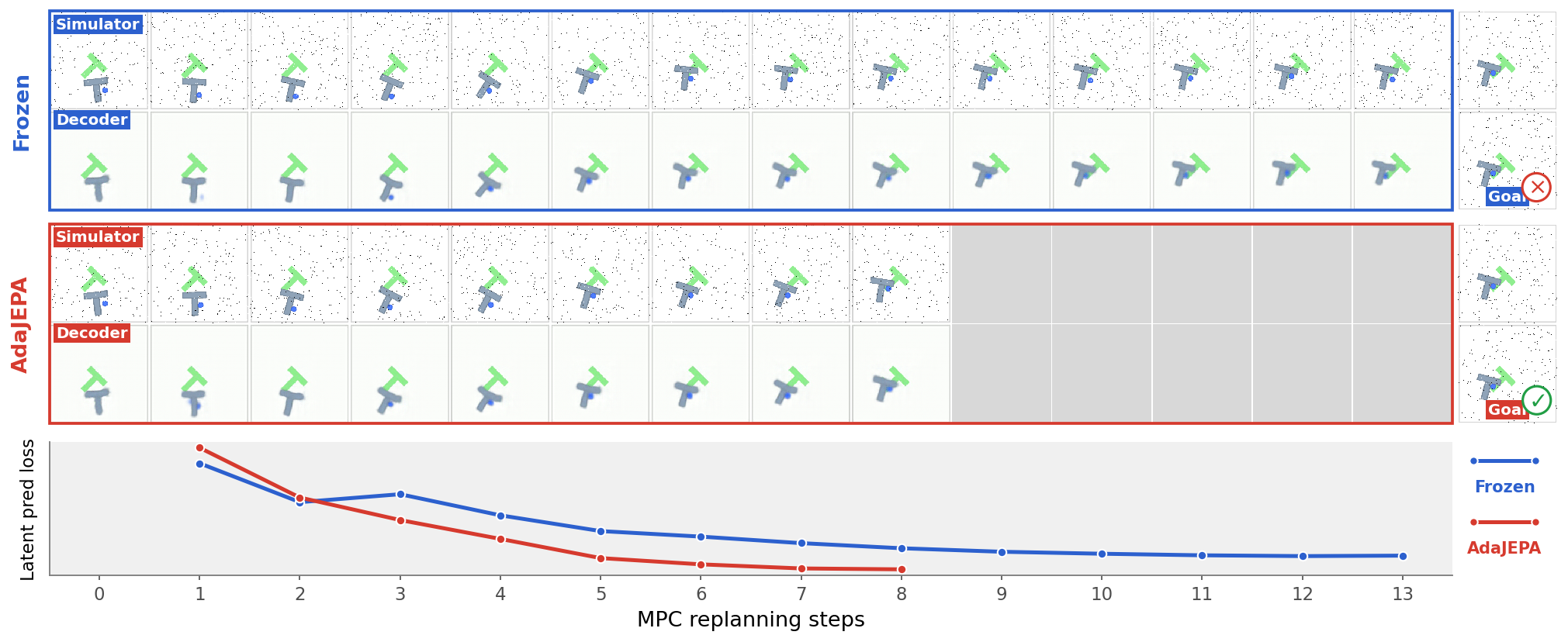}
    \vspace{-0.1in}
    \caption{Visual Shifts: The model is trained on the default PushT data, but the test-time observations have salt-and-pepper noise. AdaJEPA consistently decreases prediction loss.}
    \label{fig:example}
\end{figure}

\begin{figure}[h]
    \centering
    \includegraphics[width=\textwidth]{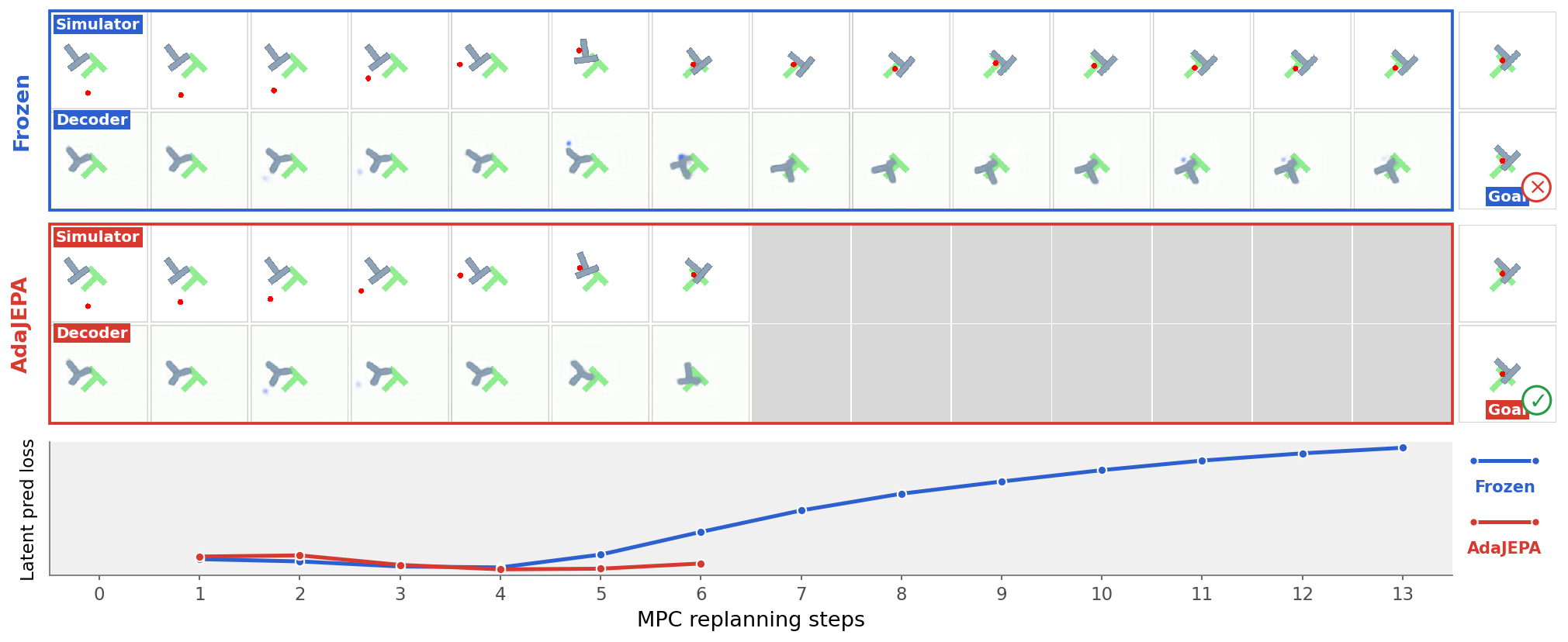}
    \vspace{-0.1in}
    \caption{Visual Shifts: The model is trained on the default PushT data (agent is blue), but the agent is red at test time. AdaJEPA consistently decreases prediction loss.}
    \label{fig:example}
    \vspace{-0.1in}
\end{figure}

\end{document}